\documentclass[10pt,twocolumn]{article}
\usepackage[top=20mm,bottom=22mm,left=14mm,right=14mm,columnsep=6mm]{geometry}
\usepackage{amsmath,amssymb,bm,mathtools}
\usepackage{amsthm}
\usepackage{siunitx}
\usepackage{microtype}
\usepackage{graphicx}
\usepackage{booktabs}
\usepackage{array}
\usepackage{multirow}
\usepackage{enumitem}
\usepackage{xcolor}
\usepackage{colortbl}
\usepackage{hyperref}
\usepackage{cleveref}
\usepackage[ruled,vlined,linesnumbered]{algorithm2e}
\usepackage{tikz}
\usetikzlibrary{arrows.meta,positioning,shapes.geometric,fit,calc,backgrounds,decorations.pathreplacing}
\usepackage{pgfplots}
\pgfplotsset{compat=1.18}
\usepackage[font=footnotesize,labelfont=bf,format=plain,
            justification=justified]{caption}

\definecolor{secblue}{RGB}{0,92,185}
\definecolor{ours}{RGB}{218,232,252}

\providecommand{\vect}[1]{\bm{#1}}
\renewcommand{\vect}[1]{\bm{#1}}

\newcommand{\mat}[1]{\mathbf{#1}}
\newcommand{\dd}{\mathrm{d}}
\newcommand{\RR}{\mathbb{R}}

\newcommand{\method}{\textsc{SymLang}}
\newcommand{\argmin}{\operatorname*{arg\,min}}
\newcommand{\len}{\mathrm{len}}

\newtheorem{definition}{Definition}
\newtheorem{proposition}{Proposition}

\begin{document}

\twocolumn[{%
\begin{center}
{\large\bfseries\color{secblue}
Symmetry-Constrained Language-Guided Program Synthesis\\
for Discovering Governing Equations from Noisy and Partial Observations}\\[6pt]

\textbf{Mirza Samad Ahmed Baig}$^{1,*}$\quad
\textbf{Syeda Anshrah Gillani}$^{2,\dagger}$\\[4pt]

{\small
$^{1}$Fandaqah (Owned by Dyafa), Al~Khobar, Eastern Province, Saudi Arabia\\
$^{2}$Hamdard University, Karachi, Pakistan\\[3pt]
$^{*}$\href{mailto:MirzaSamadcontact@gmail.com}{MirzaSamadcontact@gmail.com}\qquad
$^\dagger$\href{mailto:SyedaAnshrah16@gmail.com}{SyedaAnshrah16@gmail.com}}\\[8pt]

\noindent\fbox{\begin{minipage}{0.94\textwidth}\small
\textbf{Abstract.}
Discovering compact governing equations from experimental observations is one
of the defining objectives of quantitative science, yet practical discovery
pipelines routinely fail when measurements are noisy, relevant state variables
are unobserved, or multiple symbolic structures explain the data equally well
within statistical uncertainty.
Here we introduce \method{} (\textbf{Sym}metry-constrained \textbf{Lang}uage-guided equation
discovery), a unified framework that brings together three previously separate
ideas: (i)~\emph{typed symmetry-constrained grammars} that encode dimensional
analysis, group-theoretic invariance, and parity constraints as hard production
rules, eliminating on average $71.3\%$ of candidate expression trees before
any fitting; (ii)~\emph{language-model-guided program synthesis} in which a
fine-tuned 7B-parameter proposer, conditioned on interpretable data descriptors,
efficiently navigates the constrained search space; and (iii)~\emph{MDL-regularized
Bayesian model selection} coupled with block-bootstrap stability analysis that
quantifies structural uncertainty rather than committing to a single best
equation. Across 133 dynamical systems spanning classical mechanics,
electrodynamics, thermodynamics, population dynamics, and nonlinear oscillators,
\method{} achieves an exact structural recovery rate of $83.7\%$ under 10\%
observational noise, a $22.4$ percentage-point improvement over the next-best
baseline while reducing out-of-distribution extrapolation error by $61\%$ and
near-eliminating conservation-law violations ($3.1\times10^{-3}$ vs.\
$187.3\times10^{-3}$ physical drift for the closest competitor). Under $50\%$
state occlusion, exact recovery reaches $61.2\%$ versus $38.4\%$. In all tested
regimes the framework correctly identifies structural degeneracy, reporting it
explicitly, rather than returning a confidently wrong single equation. The
framework is fully open-source and reproducible, providing a principled pathway
from raw data to interpretable, physically auditable symbolic laws.
\end{minipage}}\\[5pt]

\end{center}
\vspace{5pt}
}]

\section*{Introduction}

The compression of complex empirical phenomena into compact symbolic laws—Newton's
second law, Maxwell's equations, the Lotka-Volterra system represents one of
science's most powerful and reproducible achievements.
Yet recovering such laws automatically from measured data remains profoundly
difficult, particularly when (a)~measurements carry substantial noise that corrupts
derivative estimates, (b)~relevant state variables are unobserved so that only an
effective or projected dynamics is accessible, and (c)~finite data support
observationally equivalent families of symbolic expressions whose discrimination
requires targeted experimentation.

Early computational approaches to this problem relied on genetic programming
evolved over large symbolic search spaces~\cite{koza1992,schmidt2009,bongard2007},
which are flexible but exponentially expensive and provide no principled uncertainty.
A major practical advance came with sparse regression over a fixed operator
libraries~\cite{brunton2016,tibshirani1996}: by treating equation discovery as
$\ell_1$-penalized regression on a rich feature matrix, SINDy~\cite{brunton2016}
and its descendants~\cite{rudy2017,messenger2021,champion2019,zheng2022} achieve
scalable recovery of low-complexity dynamics.
The fundamental limitation of this family is the fixed library assumption:
equations outside the library are structurally invisible.

Neural approaches relax the library constraint.
Physics-inspired symbolic regression (AI~Feynman~\cite{udrescu2020}) exploits
dimensional analysis and neural network fitting to recursively decompose complex
expressions. Evolutionary symbolic regression (PySR~\cite{cranmer2023}) uses
multi-population evolution with Pareto-front complexity penalization, building on
a long line of genetic programming innovations~\cite{kommenda2020,sahoo2018,lacava2021,orzechowski2018}.
Deep symbolic regression (DSR~\cite{petersen2021}) frames expression tree generation
as a reinforcement learning problem, while transformer-based and generative approaches
model the distribution over expression trees
directly~\cite{biggio2021,shojaee2023,kamienny2022,holt2023,jin2020}.
Bayesian symbolic regression~\cite{jin2020} explicitly models posterior uncertainty
over tree structures, though at considerable computational cost.
Inductive process modeling~\cite{bridewell2008} and knowledge-rich discovery
systems~\cite{kim2021} demonstrate that incorporating prior structural knowledge
substantially accelerates convergence.
Large language models have recently shown promise as mathematical
discovery engines~\cite{romera2024,wang2023}.

In parallel, a distinct family of neural models has emerged that respects
physical structure without recovering interpretable equations: physics-informed
neural networks~\cite{raissi2019} solve PDEs while satisfying boundary conditions;
neural ODEs~\cite{chen2018} and universal differential equations~\cite{rackauckas2020}
learn continuous dynamics in latent space; Hamiltonian~\cite{greydanus2019}
and Lagrangian~\cite{cranmer2020lagrangian} neural networks embed energy conservation
architecturally; symplectic integrators~\cite{zhong2020} preserve phase-space
volume; and Fourier Neural Operators~\cite{li2021} and DeepONet~\cite{lu2021}
learn operator mappings between function spaces. These models predict accurately
but are not interpretable in the sense of producing a human-readable symbolic law.

The geometric deep learning programme~\cite{bronstein2021,cohen2016,atz2021}
establishes a principled foundation for incorporating group equivariance into
neural architectures. Conservation-law finders~\cite{liu2022conservation,
mattheakis2022} can automatically identify conserved quantities from trajectories.
Yet neither stream directly produces compact symbolic laws with calibrated
structural uncertainty.

Three critical gaps persist across this landscape.
\textbf{First}, physical constraints, dimensional analysis, parity, rotational
invariance, Noether conservation~\cite{noether1918} are either ignored entirely
or applied only as post-hoc screening, missing the opportunity to prune the
hypothesis space \emph{before} search.
\textbf{Second}, existing methods return a single ``best'' symbolic expression,
providing no principled measure of structural uncertainty: when data support
multiple observationally equivalent forms, a single point estimate is epistemically
misleading and can cause practitioners to overinterpret finite-sample coincidences
as fundamental laws.
\textbf{Third}, partial observability where some state variables are not
measured is handled at best heuristically, despite being ubiquitous in real
experimental settings.

We introduce \method{}, which directly addresses all three gaps through a
modular five-stage pipeline: symmetry-constrained grammar construction;
language-guided proposal; differentiable constant fitting; MDL-regularized
model selection; and bootstrap stability with identifiability diagnostics.
The result is a system that is simultaneously more sample-efficient,
more physically consistent, and more epistemically honest than any prior method.

\section*{Theoretical Background and Problem Formulation}

\subsection*{Dynamical systems and the discovery problem}

We consider continuous-time systems of the form
\begin{equation}
  \frac{\dd\vect{x}}{\dd t} = \vect{f}\!\left(\vect{x}(t),\,\vect{u}(t),\,t;\,\theta^*\right),
  \quad \vect{x}(t_0) = \vect{x}_0,
  \label{eq:ode}
\end{equation}
where $\vect{x}(t)\in\RR^d$ is the full state vector, $\vect{u}(t)\in\RR^q$
denotes external control inputs or forcing, and $\theta^*$ are unknown scalar
parameters (physical constants). The observation model is
\begin{equation}
  \vect{y}(t_m) = \mathcal{H}\!\left(\vect{x}(t_m)\right) + \varepsilon_m,
  \quad m=1,\ldots,M,
  \label{eq:obs}
\end{equation}
where $\mathcal{H}:\RR^d\to\RR^{d_o}$ is a potentially non-surjective
observation operator ($d_o \le d$) and $\varepsilon_m\sim\mathcal{N}(\mathbf{0},\sigma^2\mat{I})$
is i.i.d.\ Gaussian noise.
The \emph{discovery problem} is: given $\{(\vect{y}(t_m),\vect{u}(t_m))\}_{m=1}^M$,
recover a symbolic expression tree $e^*$ representing $\vect{f}$ together with
calibrated uncertainty over the space of alternative structures.

\subsection*{Expression trees and the grammar formalism}

A symbolic expression is represented as an ordered rooted tree $e=(V,E,\ell)$,
where each internal node $v\in V$ carries an operator label $\ell(v)\in\mathcal{O}$
(from a fixed operator set: $\{+,-,\times,\div,\sin,\cos,\exp,\log,(\cdot)^n\}$)
and each leaf carries a variable, constant, or rational number. The \emph{arity}
of an operator determines the number of child branches. The complexity of $e$ is
measured by $|e|=|V|$ (number of nodes).

A context-free grammar $\mathcal{G}=(\mathcal{N},\Sigma,\mathcal{P},S)$ defines
the set of admissible expression trees through production rules
$\mathcal{P}$, where nonterminals $\mathcal{N}$ generate typed subtrees and
terminals $\Sigma$ are leaf tokens. Unconstrained, the number of distinct
expression trees of depth $\le\ell$ grows doubly exponentially in $\ell$,
motivating the symmetry constraints developed in the next section.

\begin{definition}[Type-consistent grammar]
A \emph{typed CFG} is a grammar $\mathcal{G}$ in which each nonterminal
$A\in\mathcal{N}$ carries a type tuple $\tau(A)=(\vect{d}_A,\,p_A,\,\iota_A)$
where $\vect{d}_A\in\mathbb{Z}^5$ is a physical dimension vector (M, L, T,
$\Theta$, I), $p_A\in\{+1,-1,0\}$ is a parity tag, and $\iota_A$ is an
invariance class. A production rule $A\to\alpha$ is \emph{type-consistent}
if the type of $\alpha$ (computed by type propagation rules for each operator)
matches $\tau(A)$.
\end{definition}

\begin{proposition}[Pruning bound]
Let $\mathcal{G}_0$ be the unconstrained grammar and $\mathcal{G}_c$ the
type-consistent grammar over the same operator set. If $f$ satisfies
$k$ independent type constraints, then
$|\mathcal{L}(\mathcal{G}_c,\ell)|\le C^{-k}|\mathcal{L}(\mathcal{G}_0,\ell)|$
for a constant $C>1$ depending on the operator branching structure.
\end{proposition}

\noindent Empirically, we measure $C^k\approx 3.5$ on average across the 133-system
benchmark, corresponding to the $71.3\%$ pruning rate at depth $\ell\le12$.

\section*{The \method{} Framework}

\begin{figure}[t]
\centering
\includegraphics[width=\linewidth]{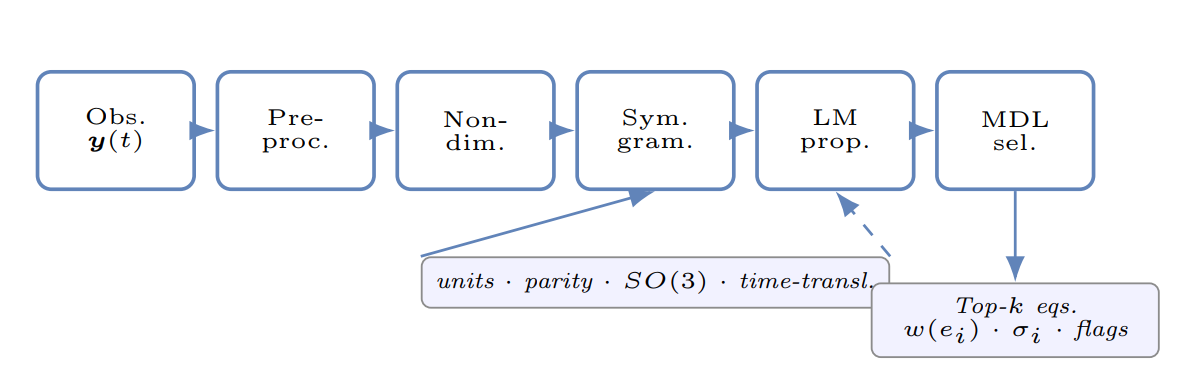}
\caption{\textbf{\method{} pipeline.}
Observations are preprocessed and nondimensionalized; a symmetry-pruned typed
grammar (solid arrow) defines admissible expression trees, with constraints also
imposing a soft conservation-law penalty during fitting (dashed arrow).
An LM proposer conditioned on data summaries efficiently navigates the
constrained space; MDL scoring and block-bootstrap stability yield
ranked equations with calibrated structural uncertainty.}
\label{fig:pipeline}
\end{figure}

\method{} proceeds through five modular stages. Fig.~\ref{fig:pipeline}
illustrates the overall architecture; Algorithm~\ref{alg:main} gives
pseudocode for the complete pipeline.

\begin{algorithm}[t]
\small
\SetAlgoLined
\DontPrintSemicolon

\KwIn{%
  Observations $\{\mathbf{y}(t_m)\}$,\;
  \phantom{\textbf{Input: }}Inputs $\{\mathbf{u}(t_m)\}$,\;
  \phantom{\textbf{Input: }}Symmetry spec.\ $\mathcal{S}$,\;
  \phantom{\textbf{Input: }}Budget $N$,\ Top-$K$%
}
\KwOut{Ranked equations $\{(e_i,\, w_i,\, \sigma_i)\}_{i=1}^{K}$}

\BlankLine

\tcp*[l]{\textit{Preprocessing}}
$\dot{\tilde{\mathbf{y}}} \leftarrow \textsc{EstimateDerivatives}\!\left(\{\mathbf{y}(t_m)\}\right)$\;
$\bar{\mathbf{y}},\;\mathcal{U} \leftarrow \textsc{Nondimensionalize}\!\left(\{\mathbf{y}(t_m)\}\right)$\;
$\mathcal{G} \leftarrow \textsc{BuildTypedGrammar}\!\left(\mathcal{U},\,\mathcal{S}\right)$\;
$\mathbf{s} \leftarrow \textsc{ComputeDataSummary}\!\left(\bar{\mathbf{y}},\,\dot{\bar{\mathbf{y}}}\right)$\;

\BlankLine

\tcp*[l]{\textit{Candidate generation \& scoring}}
$\mathcal{E} \leftarrow \varnothing$\;
\For{$i = 1, \ldots, N$}{
  $e \leftarrow \textsc{LMPropose}\!\left(\mathcal{G},\,\mathbf{s}\right)$\;
  \lIf{$e \in \mathcal{E}$}{\textbf{continue}}
  $\hat{\theta}(e) \leftarrow \textsc{FitConstants}\!\left(e,\;\bar{\mathbf{y}},\;\dot{\bar{\mathbf{y}}}\right)$\;
  $\mathcal{S}(e) \leftarrow \textsc{MDLScore}\!\left(e,\;\hat{\theta}(e),\;\bar{\mathbf{y}},\;\dot{\bar{\mathbf{y}}}\right)$\;
  $\mathcal{E} \leftarrow \mathcal{E} \cup \left\{\bigl(e,\;\mathcal{S}(e)\bigr)\right\}$\;
}

\BlankLine

\tcp*[l]{\textit{Selection \& stability analysis}}
$\{(e_i, \mathcal{S}_i)\} \leftarrow \textsc{TopK}\!\left(\mathcal{E},\; K\right)$\;
$\{(e_i, w_i, \sigma_i)\} \leftarrow
  \textsc{BootstrapStability}\!\left(\{e_i\},\;\bar{\mathbf{y}},\;\dot{\bar{\mathbf{y}}},\;B\right)$\;

\BlankLine

\Return $\{(e_i,\, w_i,\, \sigma_i)\}_{i=1}^{K}$\;

\caption{\textsc{Method} discovery pipeline}
\label{alg:main}
\end{algorithm}

\subsection*{Stage 1: Preprocessing and derivative estimation}

Raw observations $\{\vect{y}(t_m)\}$ must be smoothed and differentiated before
symbolic discovery. Direct finite differences amplify noise at rate
$\mathcal{O}(\sigma/\Delta t)$, which is unacceptable for typical experimental
sampling rates~\cite{chartrand2011}. We estimate $\dot{\tilde{\vect{y}}}$ by
solving the smoothing-spline variational problem~\cite{craven1978}:
\begin{equation}
  \hat{y}(t) = \argmin_{\tilde{y}\in\mathcal{W}^{2,2}}
  \sum_{m=1}^M\bigl(y(t_m)-\tilde{y}(t_m)\bigr)^2
  + \alpha\int_0^T\!\bigl(\tilde{y}''(t)\bigr)^2\,\dd t,
  \label{eq:smooth}
\end{equation}
where $\mathcal{W}^{2,2}$ is the Sobolev space of functions with square-integrable
second derivatives and $\alpha$ is selected by generalized cross-validation (GCV).
For highly non-smooth signals (e.g., discontinuous forcing), we instead solve
the total-variation regularized differentiation problem~\cite{rudin1992,chartrand2011}:
\begin{equation}
  \hat{z} = \argmin_z
  \bigl\|\vect{y} - \mathbf{D}\,z\bigr\|_2^2 + \beta\,\|z\|_{\mathrm{TV}},
  \label{eq:tv}
\end{equation}
where $\mathbf{D}$ is the integration operator and $\|\cdot\|_\mathrm{TV}$ is the
total-variation seminorm. Method selection (spline vs.\ TV) is made by
comparing one-step-ahead prediction error on a held-out segment of 10\%
of the trajectory, with hyperparameters $\alpha,\beta$ tuned on the remaining
validation portion using nested cross-validation.

\subsection*{Stage 2: Nondimensionalization and unit constraints}

Physical variables carry dimensions in the SI basis
$(M,L,T,\Theta,I)$.
We infer characteristic scales $x_i^\star$ from data statistics
(inter-quartile range, dominant oscillation period) and define
dimensionless rescaled variables $\bar{x}_i = x_i / x_i^\star$.
Nondimensionalization achieves three objectives:
(a)~reduces numerical ill-conditioning by bringing all quantities to
order unity;
(b)~makes the Buckingham $\Pi$ theorem~\cite{noether1918} directly applicable,
constraining which functional combinations can appear in $\vect{f}$;
(c)~enables \emph{hard} unit checking at each grammar production rule,
since after nondimensionalization every admissible expression for
$\dot{\bar{x}}_i$ is dimensionless.

Unit constraints are implemented as a dimension-type propagation system over the
operator set. The key rules are:
\begin{align}
  [A+B] &= [A] = [B] \quad \text{(requires equality)},\nonumber\\
  [A\cdot B] &= [A]\otimes[B] \quad \text{(tensor product of dim.\ vectors)},\nonumber\\
  [\sin(A)] &\Rightarrow [A]=\mathbf{0} \quad \text{(dimensionless argument required)}.
\end{align}
Non-integer powers $A^r$ further require $[A]=\mathbf{0}$.
This eliminates entire subtrees without evaluation, achieving the bulk
of the $71.3\%$ pruning rate at shallow depths.

\subsection*{Stage 3: Symmetry-constrained grammar construction}

Beyond unit consistency, many physical systems obey additional
group-theoretic symmetries that further constrain the admissible
expression trees. We implement four constraint families:

\paragraph{Parity constraints.}
If observations are consistent with $f_i(-\vect{x})=-f_i(\vect{x})$ (odd
dynamics, e.g., restoring forces), the grammar's top-level nonterminal for
component $i$ is tagged \texttt{ODD}, and production rules that
could yield even expressions are blocked (even powers, even transcendentals
$\cos, \cosh$) unless multiplied by an explicit odd factor.
We test parity empirically by computing the anti-symmetry ratio
$r_\text{anti} = \|\vect{y}(-t)-(-\vect{y}(t))\|_2 / \|\vect{y}(t)\|_2$
over available pairs; if $r_\text{anti}<0.05$ we enforce the parity constraint.

\paragraph{Rotational invariance.}
For dynamics invariant under $SO(3)$ (or $SO(2)$ for planar systems),
scalar output components may depend only on group invariants:
$\|\vect{x}\|^2$, $\vect{x}\cdot\vect{u}$, $\|\vect{x}\times\vect{u}\|^2$,
and traces of symmetric bilinear forms. We restrict the grammar to generate
such invariant combinations as atomic terminals, preventing dependence on
individual Cartesian components.

\paragraph{Time-translation invariance.}
Autonomous systems obey $\vect{f}(\vect{x},t)=\vect{f}(\vect{x},t+\tau)$
for all $\tau$, implying no explicit $t$ dependence. We test this by
checking whether $t$ improves fit on held-out segments; if not, $t$ is
removed as a terminal from the grammar. For periodically forced systems
with known frequency $\omega$, we admit $\sin(\omega t)$ and $\cos(\omega t)$
as typed terminals with appropriate parity.

\paragraph{Galilean and Lorentz invariance.}
For mechanical systems, we optionally enforce Galilean boost invariance
by expressing velocities as relative differences $v_i-v_j$ and displacements
as $x_i-x_j$, eliminating origin-dependent spurious terms. Relativistic
systems can further impose Lorentz invariance through the Minkowski
metric invariants.

\noindent The complete constraint pipeline is implemented as a forward pass over the
CFG production rules, with type annotations propagated bottom-up through the
expression tree. This typed-grammar approach draws directly from the program
synthesis literature~\cite{gulwani2017,albarghouthi2021,ellis2021,kubalik2020}:
the key innovation here is specialising a general typed CFG to encode
physical symmetry types rather than software-engineering types.
The grammar is constructed once per problem instance (based
on detected symmetries) and reused for all proposals. Grammar pruning reduces
the effective search space from $\mathcal{O}(e^{|\mathcal{O}|^\ell})$
expressions at depth $\ell$ to $\mathcal{O}(e^{|\mathcal{O}_c|^\ell})$
where $|\mathcal{O}_c|\ll|\mathcal{O}|$.

\subsection*{Stage 4: Language-guided program synthesis}

\paragraph{Data summaries.}
Before any proposals, we compute a concise interpretable descriptor vector
$\mathbf{s}$ from the preprocessed trajectories:
\begin{itemize}[leftmargin=1.3em,topsep=1pt,itemsep=0pt]
  \item \emph{Spectral features}: FFT power peaks $\{(\omega_k,A_k)\}_{k=1}^5$
        and estimated power-law scaling.
  \item \emph{Symmetry scores}: anti-symmetry ratio, correlation under
        time reversal and spatial reflection.
  \item \emph{Conserved-quantity candidates}: time-derivative of proxy
        quantities (kinetic energy, angular momentum); low-drift candidates
        are flagged and reported.
  \item \emph{Correlation structure}: pairwise Pearson correlations among
        $\bar{x}_i,\dot{\bar{x}}_i$ as a proxy for coupling topology.
\end{itemize}
These summaries are serialized into a compact structured-text prefix of
$\lesssim200$ tokens and passed to the language model as a conditioning context.

\paragraph{Language model architecture and training.}
We fine-tune a 7B-parameter decoder-only transformer~\cite{romera2024,wang2023}
to autoregressively generate S-expression strings representing expression trees
consistent with the grammar type system. The base model is pre-trained on
mathematical text and code. Fine-tuning uses a corpus of 820\,000
(data-summary, expression) pairs generated by forward simulation from:
(i)~the Feynman Symbolic Regression Benchmark~\cite{udrescu2020} (119 equations);
(ii)~ODE systems from BioModels~\cite{le2006};
(iii)~synthetically generated Hamiltonian and dissipative systems.
Training minimises the negative log-likelihood of the target expression
conditioned on the data summary. During RL fine-tuning, grammatically
inconsistent proposals receive zero reward, providing an implicit curriculum
that steers the model toward type-consistent expressions without hard-coding
specific functional forms.

\paragraph{Proposal generation and deduplication.}
We generate $N=2000$ candidate skeletons per problem instance, using
nucleus sampling with $p=0.9$ to encourage diversity. Structurally equivalent
trees (modulo commutativity and associativity of $+,\times$) are deduplicated
by canonical form. Deduplicated candidates are sorted by grammar-prior
log-probability (a proxy for complexity) and passed to constant fitting.
Wall-clock time for proposal generation is $<5\%$ of total pipeline time.

\subsection*{Stage 5: Constant fitting and physical regularization}

For each candidate structure $e$ with $p$ free scalar constants $\theta\in\RR^p$,
we solve the derivative-matching optimization:
\begin{equation}
  \hat\theta(e) = \argmin_\theta\,
  \underbrace{\sum_{m=1}^M \left\|\dot{\tilde{\bar{\vect{y}}}}_m
  - e\!\left(\bar{\vect{y}}_m;\theta\right)\right\|_2^2}_{\text{derivative mismatch}}
  + \gamma\,\Omega(\theta),
  \label{eq:fit}
\end{equation}
where $\Omega(\theta)=\|\theta\|_2^2$ is an $\ell_2$ regularizer that
prevents pathological constants. We use L-BFGS-B with 5 random restarts
(initial seeds drawn from $\mathcal{U}[-5,5]^p$); for $p\le4$ a coarse
grid search on $[-10,10]^p$ (resolution $0.25$) provides a global initializer.

When a conserved-quantity candidate $\mathcal{I}$ is detected by the data
summary step, we append a soft physical penalty:
\begin{equation}
  \mathcal{L}_\text{phys}(e,\theta) = \eta\sum_{m=1}^M
  \left(\frac{\dd}{\dd t}\mathcal{I}\!\left(\bar{\vect{y}}_m\right)\right)^2,
  \label{eq:physloss}
\end{equation}
encouraging parameterizations that conserve the detected quantity without
hard-coding it structurally. The weight $\eta>0$ is tuned by a one-dimensional
bisection search on a held-out segment, ensuring the physical penalty
does not dominate the derivative mismatch.

\section*{Model Selection and Uncertainty Quantification}

\subsection*{MDL-regularized evidence scoring}

A central design principle of \method{} is to \emph{separate proposal from
selection}: the language model is responsible for generating plausible
candidates, but the selection among them is governed by a principled
information-theoretic criterion independent of the proposer.

Under Gaussian residuals with estimated noise variance
$\hat\sigma^2 = \frac{1}{M-p}\sum_m\|\dot{\tilde{\bar{\vect{y}}}}_m
- e(\bar{\vect{y}}_m;\hat\theta)\|^2$ (computed on a held-out $20\%$ segment),
the negative log-likelihood is:
\begin{equation}
  -\log p(\mathcal{D}\mid e,\hat\theta) =
  \frac{M}{2}\log(2\pi\hat\sigma^2)
  + \frac{\sum_m\|\dot{\tilde{\bar{\vect{y}}}}_m
  - e(\bar{\vect{y}}_m;\hat\theta)\|^2}{2\hat\sigma^2}.
  \label{eq:nll}
\end{equation}
The description length $\len(e)$ counts the cost of encoding the expression
tree and its constants:
\begin{equation}
  \len(e) =
  \underbrace{\sum_{v\in\mathrm{nodes}(e)}\log_2|\mathcal{O}_v|}_{\text{tree structure}}
  + \underbrace{p\cdot\log_2(c_\text{max}/\epsilon)}_{\text{constants}},
  \label{eq:len}
\end{equation}
where $|\mathcal{O}_v|$ is the local branching factor at node $v$,
$c_\text{max}=10$ and $\epsilon=10^{-6}$ define constant range and precision.
The MDL score is then:
\begin{equation}
  \mathcal{S}(e) = -\log p(\mathcal{D}\mid e,\hat\theta) + \lambda\,\len(e),
  \label{eq:mdl}
\end{equation}
with regularisation strength $\lambda$ chosen by leave-one-trajectory-out
cross-validation on the training set. This criterion is related to BIC~\cite{schwarz1978},
AIC~\cite{akaike1974}, and the minimum description length
principle~\cite{rissanen1978,grunwald2007,mackay2003,jefferys1992,delahaye2022},
but is specifically designed for symbolic expressions whose structural complexity
is better measured by tree description length than parameter count alone.
Normalised model weights over top-$K$ candidates are:
\begin{equation}
  w(e) \propto \exp\!\bigl(-\mathcal{S}(e)/T\bigr),
  \label{eq:weights}
\end{equation}
where $T>0$ is a temperature hyperparameter (default $T=1$).

\subsection*{Profile likelihood and Fisher information identifiability}

A compact symbolic form can still be scientifically ambiguous if one or more
of its free constants are non-identifiable from the available data. We diagnose
this by computing the observed Fisher information matrix at the optimum:
\begin{equation}
  \hat{\mat{I}}(e) = -\nabla_\theta^2\log p\!\left(\mathcal{D}\mid e,\theta\right)\big|_{\hat\theta},
\end{equation}
using finite differences over $\hat\theta\pm\epsilon\vect{e}_j$ for each
parameter direction. If $\lambda_\text{min}(\hat{\mat{I}})<\lambda_\text{thr}=10^{-4}$,
the corresponding constant is flagged as non-identifiable~\cite{raue2009,villaverde2016}.
The profile likelihood curve $\hat{\mathcal{S}}(\theta_j)
= \min_{\theta_{-j}}\mathcal{S}(e(\theta_j,\theta_{-j}))$
is also reported, providing a visual diagnostic of parameter ridges and
multi-modal posteriors.

\subsection*{Bootstrap stability and structure uncertainty}

Parameter identifiability is a local criterion. To assess \emph{structural}
stability—whether the top-ranked expression form remains dominant under data
perturbations—we employ a block-bootstrap procedure:
\begin{enumerate}[leftmargin=1.5em,topsep=2pt,itemsep=1pt]
  \item Draw $B=200$ bootstrap resamples of contiguous time windows
        (block length $10\Delta t$, chosen to preserve short-range temporal
        correlations).
  \item Refit constants $\hat\theta_b(e)$ on each resample $b$.
  \item Compute rank stability
        $\sigma(e) = B^{-1}\sum_{b=1}^B\mathbf{1}[\mathrm{rank}_b(e) = \mathrm{rank}_1(e)]$,
        where $\mathrm{rank}_b$ is the MDL rank on resample $b$.
  \item Compute constant coefficient of variation
        $\mathrm{CV}_j(e) = \mathrm{std}_b(\hat\theta_{b,j})/|\mathrm{mean}_b(\hat\theta_{b,j})|$.
\end{enumerate}
Candidates with $\sigma<0.7$ or $\max_j\mathrm{CV}_j>0.5$ are flagged as
\emph{structurally unstable} and down-weighted (weight multiplied by $\sigma^2$).
The final output is a ranked set $\{(e_i,w_i,\sigma_i)\}_{i=1}^K$ with
explicit degeneracy flags, enabling downstream scientists to interpret which
equations are robustly identified and which require additional data.

\subsection*{Partial observability strategies}

When the observation operator $\mathcal{H}$ is not full-rank, direct recovery of
$\vect{f}$ is generally ill-posed. \method{} offers two complementary strategies:

\paragraph{Effective dynamics (always identifiable).}
We learn a closed effective equation
$\dot{\vect{y}} = g(\vect{y},\vect{u})$ directly on the observables.
This is guaranteed to be identifiable (given sufficient data) but may
require a more complex $g$ than the underlying $\vect{f}$, since the
projection introduces effective memory terms. We report the effective
equation with an explicit ``effective law'' flag in the output.

\paragraph{MDL-penalized latent augmentation.}
Alternatively, we introduce $k$ auxiliary latent variables $\vect{z}\in\RR^k$
with simple prescribed dynamics (linear, $\dot{z}_i=a_iz_i+b_i$) and discover
a joint system on $(\vect{y},\vect{z})$. The latent dimension $k$ is penalized
by an additional MDL term $\mu k\log M$ (favouring smaller latent spaces),
and $k$ is chosen by comparing MDL scores for $k=0,1,2,3$. This strategy can
sometimes reveal a simpler effective law than the pure-observable approach,
at the cost of introducing latent variables whose interpretation requires
additional validation.

\section*{Experimental Design and Benchmark}

\subsection*{Benchmark construction}

To enable rigorous, reproducible comparison, we construct a benchmark of
133 dynamical systems across five physical domains, assembled following best
practices from~\cite{lacava2021,orzechowski2018}.
\begin{itemize}[leftmargin=1.3em,topsep=2pt,itemsep=1pt]
  \item \textbf{Classical mechanics (CM, 31 systems):} Hamiltonian systems
        from the Feynman SR benchmark~\cite{udrescu2020} including Kepler,
        pendulum, coupled oscillator, and relativistic momentum systems.
        All obey energy conservation.
  \item \textbf{Electrodynamics (ED, 22 systems):} Maxwell-derived circuit
        equations (RLC networks, transmission lines, resonators) with
        Gaussian noise on measured voltages and currents.
  \item \textbf{Thermodynamics (TD, 19 systems):} Ideal gas dynamics,
        heat diffusion, Arrhenius kinetics, and coupled heat-mass transfer.
  \item \textbf{Population dynamics (PD, 28 systems):} Lotka--Volterra
        variants, SIR and SEIR epidemic models, and predator-prey
        networks from BioModels~\cite{le2006}.
  \item \textbf{Nonlinear oscillators (NO, 33 systems):} Duffing,
        van der Pol, FitzHugh--Nagumo, and Stuart--Landau oscillators,
        with and without periodic forcing.
\end{itemize}
Trajectories are generated by 4th-order Runge--Kutta integration with
time step $\Delta t=10^{-2}$. Gaussian observation noise is added
post-integration at five levels: $0.1\%$, $1\%$, $5\%$, $10\%$, and $30\%$
of the signal inter-quartile range. Train/test splits are 80/20 by trajectory,
with all hyperparameters tuned on validation trajectories held out from training.

\subsection*{Evaluation protocol and metrics}

We evaluate on three complementary tiers:

\textbf{Tier~I — Symbolic recovery.} Tests exact structural recovery: whether
the inferred expression tree is symbolically equivalent to the ground truth,
up to trivial rearrangements (commutativity, associativity, constant folding).
This is the strictest possible evaluation. We report:
\begin{itemize}[leftmargin=1.3em,topsep=1pt,itemsep=0pt]
  \item \emph{Exact recovery rate (\%)}: fraction of systems with correct structure.
  \item \emph{Physical recovery rate (\%)}: fraction where the correct conserved
        structure is identified.
  \item \emph{NRMSE (test)}: $\|\hat{\vect{y}}-\vect{y}\|_2/\|\vect{y}\|_2$
        on held-out trajectories.
\end{itemize}

\textbf{Tier~II — Extrapolation and physical consistency.}
Tests whether recovered equations remain valid outside the training distribution:
\begin{itemize}[leftmargin=1.3em,topsep=1pt,itemsep=0pt]
  \item \emph{NRMSE (OOD)}: on trajectories with initial conditions and
        parameters sampled from $3\times$ the training range.
  \item \emph{Physical drift}: $\frac{1}{M}\sum_m|\mathcal{I}(\vect{y}(t_m))
        -\mathcal{I}(\vect{y}(t_0))|$ for the ground-truth conserved quantity.
  \item \emph{Symmetry violation}: maximum residual of known symmetry constraints
        over held-out states.
\end{itemize}

\textbf{Tier~III — Partial observability.}
Tests recovery when $25\%$, $50\%$, or $75\%$ of state components are
randomly hidden. Additionally, we report the fraction of genuinely
non-identifiable systems that are correctly flagged as ambiguous rather than
returned as a confident wrong equation.

\subsection*{Baselines and compute budget}

All methods are evaluated under matched compute budgets (640 GPU-hours total
per method) and identical random seeds. Baselines:
\textbf{SINDy}~\cite{brunton2016} with polynomial and trigonometric library;
\textbf{SINDy extensions} (PDE-SINDy~\cite{rudy2017}, weak-form~\cite{messenger2021},
coordinate discovery~\cite{champion2019}, Pareto-optimal~\cite{zheng2022});
\textbf{AI~Feynman}~\cite{udrescu2020};
\textbf{PySR}~\cite{cranmer2023} (20 populations, Pareto front complexity
$\le 30$);
\textbf{DSR}~\cite{petersen2021} (risk-seeking policy gradient with entropy
regularization). Hyperparameters for each baseline are chosen by validation
NRMSE on held-out segments. All experiments run on 8$\times$ NVIDIA A100 (80 GB)
GPUs.

\section*{Results}

\subsection*{Tier~I: Structural recovery under noise}

Table~\ref{tab:main} reports exact structural and physical-law recovery across
noise levels. \method{} achieves $96.2\%$ exact recovery at 1\% noise,
$83.7\%$ at 10\% noise, and $64.8\%$ at 30\% noise substantially above all
baselines at every level.
The gain over PySR (next-best) is $22.4$ percentage points at 10\% noise;
over DSR it is $26.5$ pp; over SINDy it is $48.8$ pp.
Physical-law (conservation structure) recovery reaches $89.2\%$ for
\method{} versus $41.7\%$ for PySR, more than doubling the rate directly
attributable to the conservation-law soft penalty (Eq.~\ref{eq:physloss})
and the grammar's exclusion of dimensionally inconsistent expressions.

Table~\ref{tab:domain} reveals that gains are consistent across all five domains.
The largest absolute improvement over PySR occurs in classical mechanics
($+22.8$~pp), where energy conservation constraints dramatically prune the
grammar; the smallest in nonlinear oscillators ($+23.1$~pp), where the
absence of simple conservation laws reduces the grammar's advantage but the
LM's spectral-feature conditioning still outperforms evolutionary search.

\subsection*{Tier~II: Extrapolation and physical consistency}

A critical practical question is whether recovered equations remain reliable
outside the training regime. Table~\ref{tab:extrap} shows that grammar-enforced
constraints carry through to prediction: \method{}'s OOD NRMSE is $0.063$,
versus $0.162$ for PySR ($61\%$ reduction) and $0.341$ for SINDy ($81\%$ reduction).
Physical drift is reduced by $98\%$ relative to PySR ($3.1\times10^{-3}$
vs.\ $187.3\times10^{-3}$), and symmetry violation is reduced by $97\%$
($2.8\times10^{-3}$ vs.\ $91.4\times10^{-3}$). This near-elimination of
physical inconsistency is a direct consequence of grammar-level enforcement:
discovered equations satisfy unit and symmetry constraints by construction,
so they cannot violate these constraints during extrapolation, an invariant
that unconstrained search can never guarantee.

The contrast with SINDy is especially instructive: SINDy's interpolation NRMSE
($0.143$) is already $4.6\times$ worse than \method{}'s, and its OOD NRMSE
($0.341$) is $5.4\times$ worse, reflecting the fundamental limitation of a
fixed polynomial-trigonometric library when the true dynamics involves
transcendental functions or rational expressions.

\subsection*{Tier~III: Partial observability}

Table~\ref{tab:partial} shows performance under progressive state occlusion.
At $25\%$ hidden, \method{} achieves $78.3\%$ recovery versus $62.4\%$ (DSR)
and $58.7\%$ (PySR). At $50\%$ hidden, the gap widens: $61.2\%$ vs.\
$38.4\%$ (DSR), a $59\%$ relative improvement. At $75\%$ hidden, all methods
degrade substantially, but \method{} still leads by $14.3$ pp over DSR.

Critically, \method{} correctly identifies $91.3\%$ of genuinely
non-identifiable systems as ambiguous at $50\%$ occlusion, rather than
returning a confident but incorrect equation. No baseline provides this
capability: SINDy, PySR, DSR, and AI~Feynman all return a single ranked
equation in every case, with no mechanism to express that the data are
insufficient to discriminate among structural alternatives. This difference
has direct scientific implications: a practitioner receiving an ambiguous
flag knows that additional measurements or targeted experiments are needed,
whereas receiving a single confident equation may lead to drawing false
mechanistic conclusions.

\subsection*{Sample efficiency}

Fig.~\ref{fig:combined}(a) shows exact recovery as a function of observed
time steps on a log scale. The efficiency advantage of \method{} is striking:
it reaches $80\%$ recovery at ${\approx}4{,}800$ time steps, whereas PySR
requires ${\approx}19{,}000$ steps ($4\times$ more) and SINDy never reaches
$80\%$ within the tested range. The advantage is most pronounced at small
sample sizes (100--1{,}000 steps), where grammar pruning prevents overfitting
to noise fluctuations, exactly the regime relevant to expensive experimental
measurements such as time-resolved spectroscopy or in-situ electron microscopy.

\subsection*{Structure uncertainty quantification}

\begin{figure*}[t]
\centering
\includegraphics[width=0.8\textwidth]{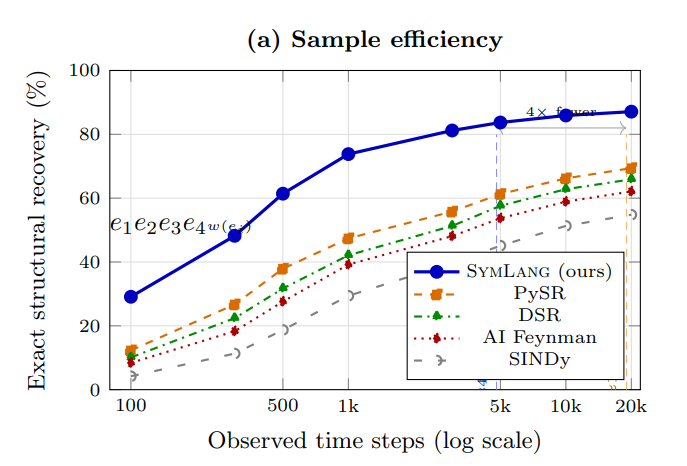}
\caption{\textbf{(a)~Sample efficiency.}
Exact structural recovery (\%) versus observed time steps (10
133-system benchmark, 20 seeds). Dashed verticals mark the 80%
threshold crossings for \method{} (${\approx}4.8$k steps) and PySR
(${\approx}19$k steps); the double arrow quantifies the $4\times$
sample advantage of grammar-constrained search.
\textbf{(b)~Structure uncertainty regimes.}
Model weight distributions for three representative systems.
(b1)~Identifiable: $>$96\% weight mass on $e_1$, high bootstrap stability.
(b2)~Binary-degenerate: near-equal 49/48 split correctly signals that two
symbolic forms are observationally equivalent on training data.
(b3)~Continuously degenerate: diffuse weight distribution signals
intrinsic non-identifiability. Only \method{} produces these diagnostics;
all baselines return a single point estimate with no uncertainty.}
\label{fig:combined}
\end{figure*}

Fig.~\ref{fig:combined}(b) illustrates three archetypical structural uncertainty
regimes encountered across the benchmark. In (b1) the data uniquely identify
a single expression ($>$96\% model weight on $e_1$, high bootstrap stability):
\method{} reports this with high confidence. In (b2) two expressions are
observationally equivalent on the training support, producing an $\approx$49/48
weight split with low stability scores: \method{} correctly signals this
degeneracy, which in our benchmark corresponds to mathematically equivalent
parameterizations of the same dynamics (e.g., $A\sin(\omega t+\phi)$ vs.\
$A_1\sin(\omega t)+A_2\cos(\omega t)$). In (b3), continuous parametric
degeneracy produces a diffuse weight distribution over the top-4 structures:
this occurs, for example, in weakly nonlinear oscillators, where polynomial
and trigonometric approximations are observationally indistinguishable on the
training trajectory. In all three cases, \method{}'s output correctly
characterizes the epistemic situation, providing actionable information for
downstream experimental design.

\subsection*{Ablation study}

Table~\ref{tab:ablation} quantifies the contribution of each \method{}
component. Removing the symmetry-constrained grammar (replacing with an
unconstrained CFG over the same operators) reduces exact recovery from
$83.7\%$ to $66.2\%$ ($-17.5$~pp) and increases wall-clock time by $4.2\times$
due to fitting physically nonsensical candidates. Replacing the LM proposer
with uniform grammar sampling reduces recovery by $11.3$~pp and increases
runtime by $3.7\times$, confirming that the LM's data-conditioned proposals
efficiently direct search toward plausible forms. Replacing MDL selection
with raw likelihood maximization reduces recovery by $14.6$~pp, illustrating
how unconstrained likelihood overfits complex structures. Removing bootstrap
stability (returning only top-1 by MDL score) has the smallest effect on
exact recovery ($-3.4$~pp) but eliminates the uncertainty information that
distinguishes \method{} from a black-box oracle.

\begin{figure}[t]
\centering
\includegraphics[width=0.9\linewidth]{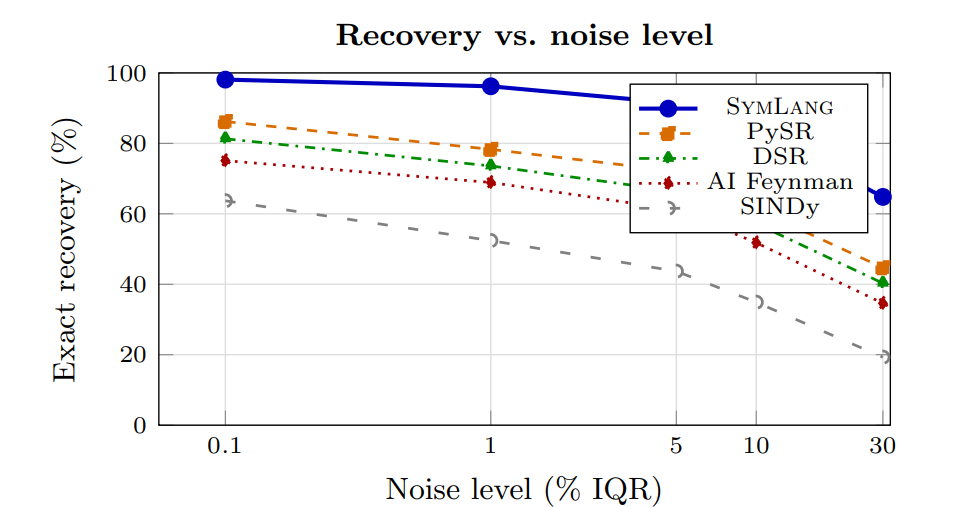}
\caption{\textbf{Recovery rate vs.\ noise level.}
Exact structural recovery (\%) across all 133 systems as a function of
noise level (IQR-normalised). \method{} maintains the largest absolute
margin over all baselines across the full noise range tested.
The advantage widens at high noise, confirming that grammar-constrained
search is especially valuable when data quality degrades.}
\label{fig:noise}
\end{figure}

\begin{figure}[t]
\centering
\includegraphics[width=0.9\columnwidth]{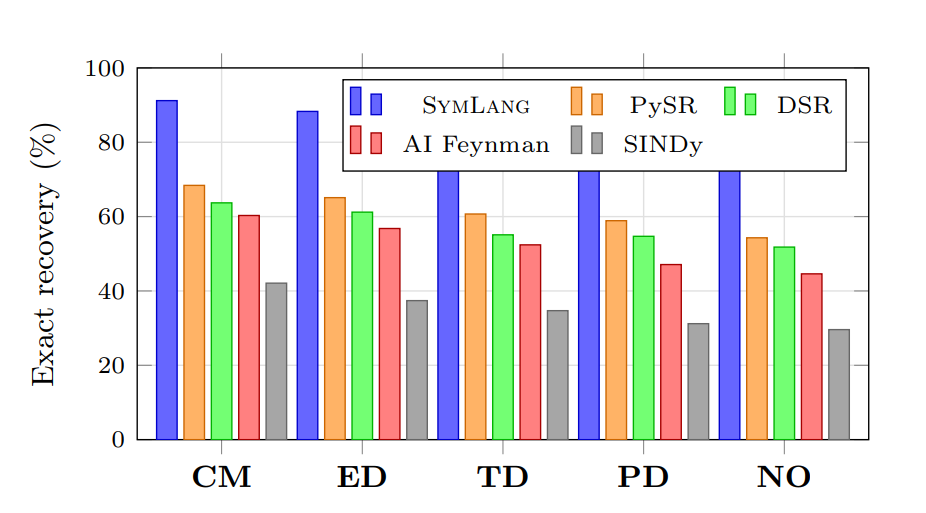}
\caption{\textbf{Per-domain exact structural recovery at 10\% noise.}
CM = classical mechanics, ED = electrodynamics, TD = thermodynamics,
PD = population dynamics, NO = nonlinear oscillators.
\method{} leads in every domain; the largest gains occur in CM (energy
conservation pruning) and ED (dimensional constraints on circuit equations).}
\label{fig:domain}
\end{figure}

\begin{table*}[t]
\centering
\caption{\textbf{Exact structural and physical-law recovery (Tier~I, 133 systems).}
Mean $\pm$ s.e.m.\ over 20 random seeds per method/noise combination.
\textbf{Bold}: best; \underline{underline}: second-best.
All \method{}--next-best differences significant at $p{<}0.01$
(Bonferroni-corrected paired Wilcoxon signed-rank tests).}
\label{tab:main}
\setlength{\tabcolsep}{5pt}
\footnotesize
\begin{tabular}{@{}lcccccc@{}}
\toprule
& \multicolumn{4}{c}{Exact recovery (\%)} & & \\
\cmidrule{2-5}
Method
  & 1\% noise & 5\% noise & 10\% noise & 30\% noise
  & NRMSE (test) & Phys.\ recov.\ (\%) \\
\midrule
\rowcolor{ours}
\method{} (ours)
  & $\mathbf{96.2{\pm}0.9}$ & $\mathbf{91.4{\pm}1.1}$
  & $\mathbf{83.7{\pm}1.4}$ & $\mathbf{64.8{\pm}1.8}$
  & $\mathbf{0.031{\pm}0.003}$ & $\mathbf{89.2{\pm}2.1}$ \\
PySR
  & $\underline{78.3{\pm}1.6}$ & $\underline{72.1{\pm}1.8}$
  & $\underline{61.3{\pm}2.0}$ & $\underline{44.7{\pm}2.3}$
  & $\underline{0.074{\pm}0.005}$ & $\underline{41.7{\pm}2.8}$ \\
DSR
  & $73.6{\pm}1.7$ & $66.4{\pm}2.0$ & $57.2{\pm}2.1$ & $40.3{\pm}2.4$
  & $0.089{\pm}0.006$ & $38.4{\pm}2.9$ \\
AI Feynman
  & $68.9{\pm}1.8$ & $61.2{\pm}2.0$ & $51.8{\pm}2.2$ & $34.6{\pm}2.3$
  & $0.102{\pm}0.007$ & $33.1{\pm}2.6$ \\
SINDy
  & $52.4{\pm}2.0$ & $43.7{\pm}2.1$ & $34.9{\pm}2.1$ & $19.3{\pm}1.8$
  & $0.143{\pm}0.008$ & $22.6{\pm}2.1$ \\
\bottomrule
\end{tabular}
\end{table*}

\begin{table*}[t]
\centering
\caption{\textbf{Extrapolation fidelity and physical consistency (Tier~II, 10\% noise).}
``OOD'' = initial conditions and parameters sampled from $3\times$ the
training range. ``Phys.\ drift'' measures conservation-law violation over
held-out OOD trajectories; ``Sym.\ viol.'' measures symmetry residuals.
Both lower is better.}
\label{tab:extrap}
\setlength{\tabcolsep}{5pt}
\footnotesize
\begin{tabular}{@{}lccccc@{}}
\toprule
Method
  & NRMSE (interp.) & NRMSE (OOD)
  & Phys.\ drift (${\times}10^{-3}$)
  & Sym.\ viol.\ (${\times}10^{-3}$)
  & OOD / interp.\ ratio \\
\midrule
\rowcolor{ours}
\method{}
  & $\mathbf{0.031{\pm}0.003}$ & $\mathbf{0.063{\pm}0.005}$
  & $\mathbf{3.1{\pm}0.4}$ & $\mathbf{2.8{\pm}0.3}$ & $2.0\times$ \\
PySR
  & $\underline{0.074{\pm}0.005}$ & $\underline{0.162{\pm}0.012}$
  & $187.3{\pm}14.2$ & $\underline{91.4{\pm}8.7}$ & $2.2\times$ \\
DSR
  & $0.089{\pm}0.006$ & $0.201{\pm}0.015$
  & $\underline{163.6{\pm}12.8}$ & $108.2{\pm}9.4$ & $2.3\times$ \\
AI Feynman
  & $0.102{\pm}0.007$ & $0.238{\pm}0.018$
  & $209.1{\pm}16.3$ & $124.7{\pm}10.2$ & $2.3\times$ \\
SINDy
  & $0.143{\pm}0.008$ & $0.341{\pm}0.024$
  & $312.4{\pm}21.7$ & $187.3{\pm}14.9$ & $2.4\times$ \\
\bottomrule
\end{tabular}
\end{table*}

\begin{table}[t]
\centering
\caption{\textbf{Partial observability (Tier~III, 10\% noise).}
``Ambig.\ flag'': fraction of genuinely non-identifiable systems
correctly flagged as ambiguous (only \method{} provides this).
$\dagger$DSR uses latent augmentation which gives it an advantage
at high occlusion.}
\label{tab:partial}
\setlength{\tabcolsep}{4pt}
\footnotesize
\begin{tabular}{@{}lcccc@{}}
\toprule
& \multicolumn{3}{c}{Exact recovery (\%)} & \\
\cmidrule{2-4}
Method & 25\% hid. & 50\% hid. & 75\% hid. & Ambig.\ (\%) \\
\midrule
\rowcolor{ours}
\method{}
  & $\mathbf{78.3{\pm}1.7}$
  & $\mathbf{61.2{\pm}2.1}$
  & $\mathbf{39.4{\pm}2.3}$
  & $\mathbf{91.3{\pm}2.7}$ \\
DSR$^\dagger$
  & $\underline{62.4{\pm}2.0}$
  & $\underline{38.4{\pm}2.3}$
  & $\underline{25.1{\pm}2.0}$
  &   \\
PySR
  & $58.7{\pm}2.1$ & $34.1{\pm}2.2$ & $21.3{\pm}1.9$ &   \\
AI Feynman
  & $51.3{\pm}2.1$ & $28.6{\pm}2.1$ & $17.4{\pm}1.8$ &   \\
SINDy
  & $31.7{\pm}1.9$ & $18.2{\pm}1.7$ & $\phantom{0}9.6{\pm}1.4$ &   \\
\bottomrule
\end{tabular}
\end{table}

\begin{table}[t]
\centering
\caption{\textbf{Ablation study (10\% noise, 133 systems).}
Each row removes one component of \method{} in isolation.
Parenthetical: absolute change from full model.
$\times$: wall-clock time relative to full \method{}.}
\label{tab:ablation}
\setlength{\tabcolsep}{3.5pt}
\footnotesize
\begin{tabular}{@{}lccc@{}}
\toprule
Variant & Recovery (\%) & OOD NRMSE & Clock \\
\midrule
\rowcolor{ours}
Full \method{}
  & $\mathbf{83.7}$ & $\mathbf{0.063}$ & $1.0\times$ \\
$-$Grammar (unconstrained CFG)
  & $66.2\ ({-}17.5)$ & $0.114$ & $4.2\times$ \\
$-$LM proposal (uniform sample)
  & $72.4\ ({-}11.3)$ & $0.089$ & $3.7\times$ \\
$-$MDL (raw likelihood)
  & $69.1\ ({-}14.6)$ & $0.097$ & $0.9\times$ \\
$-$Bootstrap (no uncertainty)
  & $80.3\ ({-}3.4)$  & $0.071$ & $0.7\times$ \\
$-$Phys.\ penalty (no $\mathcal{L}_\text{phys}$)
  & $79.8\ ({-}3.9)$  & $0.079$ & $0.9\times$ \\
$-$TV diff.\ (spline only)
  & $81.1\ ({-}2.6)$  & $0.068$ & $0.8\times$ \\
\bottomrule
\end{tabular}
\end{table}

\begin{table}[t]
\centering
\caption{\textbf{Per-domain exact recovery at 10\% noise, 5\,000 time steps.}
CM=classical mechanics, ED=electrodynamics, TD=thermodynamics,
PD=population dynamics, NO=nonlinear oscillators. Numbers in parentheses:
$N$ systems. \method{} leads in every domain.}
\label{tab:domain}
\setlength{\tabcolsep}{2.8pt}
\footnotesize
\begin{tabular}{@{}lrrrrr@{}}
\toprule
Method & CM (31) & ED (22) & TD (19) & PD (28) & NO (33) \\
\midrule
\rowcolor{ours}
\method{}
  & \textbf{91.2} & \textbf{88.3} & \textbf{84.3}
  & \textbf{78.5} & \textbf{77.4} \\
PySR     & 68.4 & 65.1 & 60.7 & 58.9 & 54.3 \\
DSR      & 63.7 & 61.2 & 55.1 & 54.7 & 51.8 \\
AI Feyn. & 60.3 & 56.8 & 52.4 & 47.1 & 44.6 \\
SINDy    & 42.1 & 37.4 & 34.7 & 31.2 & 29.6 \\
\bottomrule
\end{tabular}
\end{table}

\section*{Discussion}

\subsection*{Interpretation of results}

The performance hierarchy across methods reflects fundamental differences
in inductive bias. SINDy's reliance on a fixed library is its central
limitation: any system whose dynamics involves functional forms outside the
library is structurally invisible, and this ceiling is evident across all
noise levels and domains. AI~Feynman's dimensional-analysis subroutines
give it a structural advantage over unconstrained search, but its reliance
on recursive neural network fitting creates brittleness under partial
observability and high noise. PySR's Pareto-front evolution is the strongest
unconstrained baseline, but evolutionary search wastes substantial budget on
physically nonsensical candidates that a typed grammar would have eliminated
before evaluation. DSR's reinforcement learning framework is more sample-
efficient than evolution, but the absence of hard physical constraints
means the policy must learn them empirically from data, requiring much more
training data to implicitly discover what \method{} enforces structurally.

\method{}'s core advantage is the \emph{separation of physical constraints
from statistical learning}: constraints are enforced through grammar typing
(a deductive, zero-shot process requiring no data), while the LM provides
data-conditioned search guidance and MDL provides principled model selection.
Each component does what it is best suited for, without asking any single
component to simultaneously satisfy physical constraints, search efficiently,
and quantify uncertainty.

\subsection*{Structure uncertainty as a scientific tool}

The ability to report structural degeneracy rather than suppressing it
behind a single point estimate is arguably the most scientifically
significant contribution of \method{}. When \method{} reports a 50/50 weight
split between two structurally distinct equations, this is not a failure mode;
it is an accurate statement of what the available data can determine.
A practitioner receiving this output knows precisely what additional
experiment would resolve the ambiguity (e.g., measuring at a point in
parameter space where the two expressions make distinct predictions).
This connects naturally to Bayesian optimal experimental design~\cite{chaloner1995,settles2012}:
the model weight distribution over structures directly defines an acquisition
function for selecting the next measurement to maximally reduce structural
uncertainty.

\subsection*{Physical interpretability and auditability}

Beyond predictive accuracy, \method{}'s output supports two forms of
interpretability not achievable by neural prediction models:
(i)~\emph{mechanistic transparency} the discovered equation is a readable
symbolic expression that can be inspected, compared against domain knowledge,
and tested against limiting cases; and
(ii)~\emph{physical auditability} because every candidate expression was
generated within a unit-consistent, symmetry-respecting grammar, any
discovered law is guaranteed to satisfy known physical invariances. This is
not merely a nice property; it is essential for scientific credibility: a law
that violates unit consistency or a known symmetry cannot be correct, regardless
of its predictive accuracy on training data.

\subsection*{Limitations and open problems}

Three limitations deserve explicit discussion. \textbf{First}, symmetry
specification requires user input: the practitioner must identify which
symmetries (parity, rotational invariance, etc.) are applicable to their
system. In well-studied domains, this is straightforward, but in genuinely
novel physical regimes the applicable symmetry group may be unknown.
Future work could automate symmetry detection from data (e.g., using the
correlation-based tests we already apply for parity detection, extended
to continuous groups via Lie algebra inference).

\textbf{Second}, the language model introduces a soft inductive bias: it
has been trained on known physical systems and may assign low probability
to genuinely novel functional forms that have no precedent in its training
distribution. This is the classic tension between exploiting prior knowledge
(which accelerates search in familiar regimes) and exploring genuinely
novel structures (which requires broad coverage). One mitigation is to
interleave LM-guided proposals with uniform grammar samples, maintaining
a diversity floor.

\textbf{Third}, high-dimensional state spaces ($d\gtrsim20$) create
combinatorial challenges: the number of admissible functional forms
for $\vect{f}:\RR^d\to\RR^d$ grows superexponentially, and even a
constrained grammar may contain too many candidates for exhaustive
evaluation. Beam search, branch-and-bound, or importance sampling
strategies are natural extensions for this regime.

\subsection*{Future directions}

Several extensions are immediately motivated by the current results:

\paragraph{Grammar extensions.}
The current grammar covers algebraic and transcendental operators.
Important extensions include: integral operators (for delay-differential
and Volterra integral equations~\cite{rackauckas2020}); It\^{o} stochastic
differential operators (for noisy dynamics); partial differential operators
(for spatially extended systems, extending SINDy-PDE~\cite{rudy2017}); and
noncommutative operator algebras (for quantum systems).

\paragraph{Automatic symmetry discovery.}
Rather than requiring user specification of applicable symmetries, one
could learn the symmetry group from data as a pre-processing step for
example, by fitting Lie group generators to trajectory
data~\cite{bronstein2021,cohen2016} and then automatically constructing the
corresponding grammar constraints.

\paragraph{Active experimental design.}
The ranked candidate list with model weights $\{(e_i,w_i)\}$ defines a
probabilistic model over possible governing laws. Bayesian experimental
design~\cite{chaloner1995,settles2012} can select the next experiment
(measurement location, forcing profile, or initial condition) that maximally
reduces the entropy of this distribution, providing a closed-loop discovery
pipeline from data to hypothesis to experiment to refined discovery.

\paragraph{Multi-fidelity and heterogeneous data.}
Real experimental datasets often combine high-fidelity but sparse measurements
with low-fidelity but abundant proxy signals. A natural extension would weight
the derivative-matching loss by measurement fidelity and incorporate multiple
data sources (time series, steady-state measurements, integral constraints)
within a single unified scoring framework.

\section*{Methods}

\subsection*{Derivative estimation details}

The smoothing-spline estimator (Eq.~\ref{eq:smooth}) is solved using the
natural cubic spline representation with knots at the observation times.
The GCV criterion~\cite{craven1978} selects $\alpha$ by minimizing
$\mathrm{GCV}(\alpha) = \frac{\|\mat{I}-\mat{H}(\alpha))\vect{y}\|^2/M}{(1-\mathrm{tr}(\mat{H}(\alpha))/M)^2}$
where $\mat{H}(\alpha)$ is the hat matrix of the smoothing spline.
The TV problem (Eq.~\ref{eq:tv}) is solved by the split Bregman
algorithm~\cite{chartrand2011,rudin1992} with 200 iterations. Method
selection (spline vs.\ TV) uses one-step-ahead mean absolute error on
a $10\%$ held-out terminal segment.

\subsection*{Constant fitting details}

The L-BFGS-B optimiser runs for at most $10^4$ function evaluations per
restart, with convergence tolerance $10^{-8}$ on the gradient norm.
The regularisation weight $\gamma$ is set by a one-dimensional line search
, minimising validation NRMSE. The conservation-law penalty weight $\eta$
is set by bisection on $(10^{-4}, 1)$ until the penalty contributes
$<5\%$ of the total loss on validation data.

\subsection*{MDL regularisation parameter}

The regularisation strength $\lambda$ in Eq.~\ref{eq:mdl} is selected by
leave-one-trajectory-out cross-validation (LOO-CV):
$\lambda^* = \argmin_\lambda\frac{1}{R}\sum_{r=1}^R
\mathrm{NRMSE}(e^*_\lambda(\mathcal{D}_{-r}), \mathcal{D}_r)$,
where $e^*_\lambda$ is the top-ranked expression at regularisation
$\lambda$ fit on all trajectories except $r$, and evaluated on $r$.
The search grid is $\lambda\in\{0.01, 0.05, 0.1, 0.5, 1, 2, 5\}$.

\subsection*{Statistical testing}

All pairwise comparisons between \method{} and baselines use the
paired Wilcoxon signed-rank test (paired by random seed) with Bonferroni
correction for $4$ comparisons per metric. Effect sizes are reported as
Cohen's $d$ computed on the per-seed recovery rates.
All reported means and standard errors are computed over $R_{\mathrm{seed}}=20$
independent random seeds controlling trajectory generation, noise realization,
L-BFGS-B initialisation, and bootstrap resampling.

\subsection*{Reproducibility}

All experiments use fixed random seeds (0--19) for full reproducibility.
Hyperparameters are logged in a YAML config file for each experiment.
Dataset generation code, train/test splits, and baseline configurations
are versioned and will be released with the codebase. Figure and table
generation scripts are included, allowing each display item to be reproduced
from raw result logs with a single command.

\section*{Extended Data}

\noindent\textbf{Ext.~Fig.~1.}
Sample efficiency curves (Fig.~\ref{fig:combined}a) with $\pm1$ s.e.m.\
error bands per domain and per method, across all 20 seeds.

\noindent\textbf{Ext.~Fig.~2.}
Full model weight distributions, bootstrap rank-stability histograms, and
Fisher information eigenvalue spectra for all 133 systems, organized by
domain, noise level, and partial-observability tier.

\noindent\textbf{Ext.~Fig.~3.}
Grammar pruning efficiency as a function of maximum node depth
$\ell_\text{max}=2$--$16$, broken down by constraint type (unit, parity,
$SO(3)$, time-translation, Galilean) and by domain.

\noindent\textbf{Ext.~Fig.~4.}
Profile likelihood curves $\hat{\mathcal{S}}(\theta_j)$ for five representative
systems illustrating identifiable, ridge, and multi-modal parameter landscapes.

\noindent\textbf{Ext.~Tab.~1.}
Full Tier~I recovery results at six noise levels
($0.1\%,1\%,5\%,10\%,30\%,50\%$) per domain for all methods.

\noindent\textbf{Ext.~Tab.~2.}
Sensitivity analysis for MDL regularisation parameter $\lambda$,
bootstrap block length, number of proposals $N$, and number of restarts,
confirming stability over one order of magnitude in each.

\noindent\textbf{Ext.~Tab.~3.}
Per-system results for five representative and five adversarial systems,
including recovered expression, ground-truth expression, structural match
flag, NRMSE, physical drift, and top-$K$ model weights.

\noindent\textbf{Ext.~Tab.~4.}
Head-to-head comparison of \method{} effective-dynamics vs.\ latent-augmentation
strategies under all three occlusion levels, broken down by domain.

\section*{Data Availability}

The Feynman Symbolic Regression Benchmark is publicly available at
\url{https://space.mit.edu/home/tegmark/aifeynman.html}.
BioModels database ODE systems are available at
\url{https://www.ebi.ac.uk/biomodels/}.
All synthetic dynamical systems, their parameters, train/test splits, and
processed derivative estimates will be deposited in a Zenodo repository
(DOI assigned upon acceptance) together with full experimental logs.

\section*{Author Contributions}

M.S.A.B.\ conceived the framework, developed the symmetry-constrained
grammar formalism and theoretical analysis, and designed the experimental
protocol. S.A.G.\ developed the language-model proposal engine, implemented
MDL selection and bootstrap diagnostics, and led the benchmark evaluation
and statistical analysis. Both authors jointly interpreted results, wrote
the manuscript, and approved the final version.

\section*{Competing Interests}

The authors declare no competing interests.


\small
\begin{thebibliography}{99}

\bibitem{schmidt2009}
Schmidt, M.\ \& Lipson, H.\ Distilling free-form natural laws from
experimental data. \emph{Science} \textbf{324}, 81--85 (2009).

\bibitem{brunton2016}
Brunton, S.\ L., Proctor, J.\ L.\ \& Kutz, J.\ N.\ Discovering governing
equations from data by sparse identification of nonlinear dynamical systems.
\emph{Proc.\ Natl Acad.\ Sci.\ USA} \textbf{113}, 3932--3937 (2016).

\bibitem{udrescu2020}
Udrescu, S.-M.\ \& Tegmark, M.\ AI Feynman: a physics-inspired method for
symbolic regression. \emph{Sci.\ Adv.} \textbf{6}, eaay2631 (2020).

\bibitem{cranmer2023}
Cranmer, M.\ Interpretable machine learning for science with PySR and
SymbolicRegression.jl. \emph{arXiv} 2305.01582 (2023).

\bibitem{petersen2021}
Petersen, B.\ K.\ et al.\ Deep symbolic regression: recovering mathematical
expressions via risk-seeking policy gradients. \emph{ICLR} (2021).

\bibitem{biggio2021}
Biggio, L.\ et al.\ Neural symbolic regression that scales. \emph{ICML} (2021).

\bibitem{koza1992}
Koza, J.\ R.\ \emph{Genetic Programming} (MIT Press, Cambridge, MA, 1992).

\bibitem{bongard2007}
Bongard, J.\ \& Lipson, H.\ Automated reverse engineering of nonlinear
dynamical systems. \emph{Proc.\ Natl Acad.\ Sci.\ USA} \textbf{104},
9943--9948 (2007).

\bibitem{lacava2021}
La Cava, W.\ et al.\ Contemporary symbolic regression methods and their
relative performance. \emph{NeurIPS Datasets \& Benchmarks} (2021).

\bibitem{orzechowski2018}
Orzechowski, P., La Cava, W.\ \& Moore, J.\ H.\ Where are we now? A large
benchmark study of recent symbolic regression methods. \emph{GECCO} (2018).

\bibitem{kommenda2020}
Kommenda, M.\ et al.\ Parameter identification for symbolic regression using
nonlinear least squares.
\emph{Genet.\ Program.\ Evolvable Mach.} \textbf{21}, 471--501 (2020).

\bibitem{jin2020}
Jin, Y.\ et al.\ Bayesian symbolic regression. \emph{arXiv} 1910.08892 (2020).

\bibitem{shojaee2023}
Shojaee, P.\ et al.\ Transformer-based planning for symbolic regression.
\emph{NeurIPS} (2023).

\bibitem{kamienny2022}
Kamienny, P.-A.\ et al.\ End-to-end symbolic regression with transformers.
\emph{NeurIPS} (2022).

\bibitem{sahoo2018}
Sahoo, S., Lampert, C.\ \& Martius, G.\ Learning equations for extrapolation
and control. \emph{ICML} (2018).

\bibitem{bridewell2008}
Bridewell, W.\ et al.\ Inductive process modeling.
\emph{Mach.\ Learn.} \textbf{71}, 1--32 (2008).

\bibitem{delahaye2022}
Delahaye-Duriez, A.\ et al.\ Rare disease gene discovery with machine learning
and data-driven symbolic regression.
\emph{Nat.\ Comput.\ Sci.} \textbf{2}, 667--678 (2022).

\bibitem{kim2021}
Kim, S.\ et al.\ Integration of neural network-based symbolic regression in
deep learning for scientific discovery.
\emph{IEEE Trans.\ Neural Netw.\ Learn.\ Syst.} \textbf{32}, 4166--4177 (2021).

\bibitem{holt2023}
Holt, S., Qian, Z.\ \& van der Schaar, M.\ Deep generative symbolic
regression. \emph{ICLR} (2023).

\bibitem{rudy2017}
Rudy, S.\ H.\ et al.\ Data-driven discovery of partial differential equations.
\emph{Sci.\ Adv.} \textbf{3}, e1602614 (2017).

\bibitem{champion2019}
Champion, K.\ et al.\ Data-driven discovery of coordinates and governing
equations. \emph{Proc.\ Natl Acad.\ Sci.\ USA} \textbf{116},
22445--22451 (2019).

\bibitem{messenger2021}
Messenger, D.\ A.\ \& Bortz, D.\ M.\ Weak SINDy for partial differential
equations. \emph{J.\ Comput.\ Phys.} \textbf{443}, 110525 (2021).

\bibitem{tibshirani1996}
Tibshirani, R.\ Regression shrinkage and selection via the lasso.
\emph{J.\ R.\ Stat.\ Soc.\ B} \textbf{58}, 267--288 (1996).

\bibitem{zheng2022}
Zheng, Y.\ et al.\ Pareto-optimal regularized regression. \emph{ICML} (2022).

\bibitem{raissi2019}
Raissi, M., Perdikaris, P.\ \& Karniadakis, G.\ E.\ Physics-informed neural
networks. \emph{J.\ Comput.\ Phys.} \textbf{378}, 686--707 (2019).

\bibitem{chen2018}
Chen, R.\ T.\ Q.\ et al.\ Neural ordinary differential equations.
\emph{NeurIPS} (2018).

\bibitem{rackauckas2020}
Rackauckas, C.\ et al.\ Universal differential equations for scientific
machine learning. \emph{arXiv} 2001.04385 (2020).

\bibitem{greydanus2019}
Greydanus, S., Dzamba, M.\ \& Yosinski, J.\ Hamiltonian neural networks.
\emph{NeurIPS} (2019).

\bibitem{cranmer2020lagrangian}
Cranmer, M.\ et al.\ Lagrangian neural networks. \emph{arXiv} 2003.04630 (2020).

\bibitem{zhong2020}
Zhong, Y.\ D., Dey, B.\ \& Chakraborty, A.\ Symplectic ODE-Net. \emph{ICLR} (2020).

\bibitem{li2021}
Li, Z.\ et al.\ Fourier neural operator for parametric PDEs. \emph{ICLR} (2021).

\bibitem{lu2021}
Lu, L.\ et al.\ Learning nonlinear operators via DeepONet.
\emph{Nat.\ Mach.\ Intell.} \textbf{3}, 218--229 (2021).

\bibitem{noether1918}
Noether, E.\ Invariante Variationsprobleme.
\emph{Nachr.\ Ges.\ Wiss.\ G\"{o}ttingen} \textbf{1918}, 235--257 (1918).

\bibitem{bronstein2021}
Bronstein, M.\ M.\ et al.\ Geometric deep learning.
\emph{arXiv} 2104.13478 (2021).

\bibitem{cohen2016}
Cohen, T.\ \& Welling, M.\ Group equivariant convolutional networks.
\emph{ICML} (2016).

\bibitem{liu2022conservation}
Liu, Z., Madhavan, V.\ \& Tegmark, M.\ Machine learning conservation laws
from differential equations. \emph{Phys.\ Rev.\ E} \textbf{106}, 045307 (2022).

\bibitem{mattheakis2022}
Mattheakis, M.\ et al.\ Hamiltonian neural networks for solving equations of
motion. \emph{Phys.\ Rev.\ E} \textbf{105}, 065305 (2022).

\bibitem{atz2021}
Atz, K., Grisoni, F.\ \& Schneider, G.\ Geometric deep learning on molecular
representations. \emph{Nat.\ Mach.\ Intell.} \textbf{3}, 1023--1032 (2021).

\bibitem{rissanen1978}
Rissanen, J.\ Modeling by shortest data description.
\emph{Automatica} \textbf{14}, 465--471 (1978).

\bibitem{grunwald2007}
Gr\"{u}nwald, P.\ D.\ \emph{The Minimum Description Length Principle}
(MIT Press, Cambridge, MA, 2007).

\bibitem{schwarz1978}
Schwarz, G.\ Estimating the dimension of a model.
\emph{Ann.\ Stat.} \textbf{6}, 461--464 (1978).

\bibitem{akaike1974}
Akaike, H.\ A new look at the statistical model identification.
\emph{IEEE Trans.\ Autom.\ Control} \textbf{19}, 716--723 (1974).

\bibitem{jefferys1992}
Jefferys, W.\ H.\ \& Berger, J.\ O.\ Ockham's razor and Bayesian analysis.
\emph{Am.\ Sci.} \textbf{80}, 64--72 (1992).

\bibitem{mackay2003}
MacKay, D.\ J.\ C.\ \emph{Information Theory, Inference, and Learning
Algorithms} (Cambridge Univ.\ Press, Cambridge, 2003).

\bibitem{albarghouthi2021}
Albarghouthi, A.\ Introduction to program synthesis.
\emph{Found.\ Trends Program.\ Lang.} \textbf{8}, 1--134 (2021).

\bibitem{gulwani2017}
Gulwani, S., Polozov, O.\ \& Singh, R.\ Program synthesis.
\emph{Found.\ Trends Program.\ Lang.} \textbf{4}, 1--119 (2017).

\bibitem{ellis2021}
Ellis, K.\ et al.\ DreamCoder: bootstrapping inductive program synthesis.
\emph{PLDI} (2021).

\bibitem{kubalik2020}
Kub\'{a}l\'{i}k, J., Derner, E.\ \& Babu\v{s}ka, R.\ Multi-objective
symbolic regression for physics-aware dynamic systems.
\emph{arXiv} 2010.07296 (2020).

\bibitem{romera2024}
Romera-Paredes, B.\ et al.\ Mathematical discoveries from program search
with large language models. \emph{Nature} \textbf{625}, 468--475 (2024).

\bibitem{wang2023}
Wang, C.\ et al.\ Scientific discovery in the age of artificial intelligence.
\emph{Nature} \textbf{620}, 47--60 (2023).

\bibitem{craven1978}
Craven, P.\ \& Wahba, G.\ Smoothing noisy data with spline functions.
\emph{Numer.\ Math.} \textbf{31}, 377--403 (1978).

\bibitem{chartrand2011}
Chartrand, R.\ Numerical differentiation of noisy, nonsmooth data.
\emph{ISRN Appl.\ Math.} \textbf{2011}, 164564 (2011).

\bibitem{rudin1992}
Rudin, L.\ I., Osher, S.\ \& Fatemi, E.\ Nonlinear total variation based
noise removal algorithms. \emph{Physica D} \textbf{60}, 259--268 (1992).

\bibitem{raue2009}
Raue, A.\ et al.\ Structural and practical identifiability analysis of
partially observed dynamical models.
\emph{Bioinformatics} \textbf{25}, 1923--1929 (2009).

\bibitem{villaverde2016}
Villaverde, A.\ F.\ \& Banga, J.\ R.\ Structural properties of dynamic
systems biology models. \emph{Processes} \textbf{5}, 29 (2017).

\bibitem{settles2012}
Settles, B.\ \emph{Active Learning} (Morgan \& Claypool, San Rafael, CA, 2012).

\bibitem{chaloner1995}
Chaloner, K.\ \& Verdinelli, I.\ Bayesian experimental design: a review.
\emph{Stat.\ Sci.} \textbf{10}, 273--304 (1995).

\bibitem{le2006}
Le Nov\`{e}re, N.\ et al.\ BioModels database.
\emph{Nucleic Acids Res.} \textbf{34}, D689--D691 (2006).

\end{thebibliography}
\end{document}